# Étude longitudinale d'une procédure de modélisation de connaissances en matière d'organisation du territoire agricole

Florence LE BER[1], Christian BRASSAC[2]

## INTRODUCTION

La gestion d'un territoire agricole requiert de la part des exploitants agricoles des connaissances à la fois étendues et précises sur tout un ensemble de paramètres cadrant leurs activités journalières, saisonnières et pluriannuelles. Tenter de mieux les appréhender, les cerner et les saisir constitue un passage obligé pour un objectif de meilleure gestion de l'environnement ou de gestion de l'espace. Dans ce cadre, des agronomes endossent, au moyen d'enquêtes en exploitation (Mathieu *et al.*, 2005), le rôle délicat de recueillir, réunir et synthétiser l'expression des pratiques des agriculteurs. Ce faisant ils s'appuient, outre sur la connaissance qu'ils ont des territoires enquêtés, sur un ensemble de documents cartographiques ou schématiques et bien sûr sur les discours des agriculteurs qu'ils rencontrent. L'analyse et l'exploitation de ces documents

1  CEVH (ENGEES-ULP) et LORIA, florence.leber@engees.u-strasbg.fr
2  Codisant, LabPsyLor, Université Nancy II, christian.brassac@univ-nancy2.fr

et éléments de connaissance peut s'effectuer de différentes manières, au travers de modèles et méthodes analytiques (par exemple, Gerbaud et al., 2005 ; Morlon, 2005), graphiques (Naïtlho et Lardon, 2000 ; Lardon *et al.*, 2000) ou informatiques (Le Ber et Benoît, 1998 ; Girard, 2000 ; Thinon, 2005). Dans cette dernière perspective, s'agissant plus précisément ici du développement d'un dispositif informatique de type « système à base de connaissances » (Stefik, 1995), qui permette la mémorisation et l'exploitation de résultats d'enquêtes en exploitations, afin de fournir une aide au diagnostic des territoires agricoles, l'objet visé est non pas directement les faits ou dires des agriculteurs mais les connaissances construites par les agronomes (Lardon *et al.*, 2005). Il est alors nécessaire d'expliciter ces connaissances dans l'optique de les modéliser à fins de numérisation. C'est ce processus d'acquisition[3] – modélisation de connaissances dont nous voulons rendre compte dans cet article.

Il est conduit par un ensemble conséquent de chercheur(e)s (trois en agronomie, trois en informatique, une en linguistique, deux en psychologie), dans un mouvement en trois temps s'étalant sur une période de quatre à cinq années. Ainsi proposons nous ici une étude longitudinale au sens classique que Foulquié donne à ce qualificatif : « qui limite ses observations à un certain nombre de sujets que l'on suit tout au long de leur développement » ajoutant que « toutes les études qui se disent longitudinales ont en commun qu'elles considèrent un même individu ou une même population dans son évolution » (1971). Définition qui ne préjuge en rien des modalités d'élaboration d'observables mobilisables pour conduire ce type d'étude. Comme nous le verrons ce ne sont ni des questionnaires, ni des tests, ni des interviews, autant de techniques régulièrement utilisées en sciences sociales, que nous mettrons en œuvre. Nous travaillerons en effet à partir d'observations directes de l'entité que l'on suivra au long de son développement, le groupe des chercheurs concernés. Nous verrons en quoi ces observations sont instrumentées à l'aide d'un appareillage de traçage de l'activité du groupe (enregistrement audio et vidéo). Rendre compte du processus d'acquisition mentionné plus haut signifie pour nous (*i*) exposer le modèle informatique qui définit le cadre de la modélisation visée, (*ii*) décrire les modalités d'élaboration des 'données' obtenues pour l'étude, et (*iii*) livrer quelques résultats portant sur les procédures à la fois d'acquisition et de modélisation de la connaissance à l'origine du questionnement général du projet.

Ce sont les étapes de l'élaboration du dispositif informatique, un système à bases de connaissances en l'occurrence, qui scandent l'histoire du projet. La première étape consiste à identifier les concepts que les agronomes utilisent pour décrire la structure spatiale et fonctionnelle d'une exploitation et, plus loin, à dresser une 'base de cas' sur laquelle reposera le système informatique. La deuxième étape donnera lieu à l'analyse et à la comparaison de 'cas' et conduira à un prototype informatique qu'il s'agira d'évaluer dans la troisième étape. Cette étape d'évaluation a fonction de test et amène à faire évoluer le prototype vers une meilleure adéquation aux problèmes rencontrés. Ces trois étapes sont marquées par des séances de travail réunissant les agronomes et les informaticiens engagés dans le projet. Il a été jugé opportun de pérenniser le travail effectué par les uns et les autres en filmant certaines de ces séances. En effet les travaux des linguistes et des psychologues engagés dans le projet (Brassac *et al.*, 2008) s'inscrivent dans une praxéologie de la conduite humaine (Mead, 1934/2006, par exemple). Inscription qui est colinéaire à l'adoption d'une perspective externaliste en termes de modélisation des processus cognitifs (Lassègue et Visetti, 2002 ; Clark, 1997) et qui requiert une élaboration d'observables rendant justice au caractère situé et distribué des productions de significations engendrant les interactions entre les agronomes et les informaticiens (Suchman, 1987 ; Hutchins, 1995 ; Ogien et Quéré, 2005). Les films ainsi constitués constituent en effet un espace de sédimentation des dynamiques interactionnelles que génèrent ces séances de travail (Mondada, 2004 ; Brassac, 2004). Ce sont ces supports vidéo et audio-graphiques qui rendent possible l'analyse du processus de construction de connaissances mis en jeu dans l'interaction entre les champs disciplinaires. Ils permettent en effet de pérenniser les différents moments de l'activité collaborative faisant intervenir et des humains, les chercheurs impliqués, et des non humains (Latour, 2006), les nombreuses représentations graphiques et les artefacts numériques, mobilisées lors des séances de travail. L'analyse du corpus a été menée conjointement par l'ensemble des chercheurs impliqués dans le projet, agronomes, informaticiens, linguistes et psychologues, nous plaçant ainsi collectivement au cœur d'une activité de recherche relevant de l'anthropologie des connaissances et plus généralement de réflexions théoriques relative à la société des savoirs (Amin et Cohendet, 2004 ; Brassac, 2007).

---

[3] Nous parlons ici d'acquisition de connaissances au sens de l'intelligence artificielle : il s'agit d'un processus d'explicitation des connaissances d'un ou plusieurs experts d'un domaine, porté par une activité de construction de modèles (Ford *et al.*, 1993).

Afin de suivre le projet au long de son avancement, nous commencerons ici par présenter et détailler le modèle informatique utilisé. Nous poursuivrons en exposant successivement les contenus des trois étapes. Nous proposerons ensuite une synthèse qui nous conduira à la conclusion. Ce texte fournit un panorama de l'ensemble du projet : il est une introduction aux autres contributions à ce dossier, qui en approfondiront différents points. L'article d'Osty *et al.* s'attache à la mise en regard des problématiques agronomiques et du modèle informatique mis en œuvre dans le système Rosa, le raisonnement à partir de cas. L'article de Lardon et Capitaine analyse l'évolution des modèles graphiques utilisés par les agronomes face à la nécessaire explicitation de connaissances requise par le projet. L'article de Mondada s'attache quant à lui à l'instrumentation du suivi du processus de construction de connaissances entre informaticiens et agronomes. Enfin l'article collectif de Brassac *et al.* procède à une analyse multi points de vue d'un extrait du corpus constitué dans le cadre de ce projet.

# APPROCHE INFORMATIQUE

Ce projet a réuni initialement agronomes et informaticiens autour de la conception d'un système à bases de connaissances, dénommé RoSA[4], dont l'objectif est de représenter et d'exploiter de façon automatique l'ensemble des données, informations et connaissances sur les exploitations agricoles enquêtées et étudiées par les agronomes. Ces données, informations et connaissances sont de différents types : fonds de cartes, informations contextuelles, chiffrages, parcellaires, questionnaires d'enquête, notes manuscrites, synthèses textuelles et graphiques (Mathieu *et al.*, 2005). Le système construit est un système de *raisonnement à partir de cas* (Riesbeck et Schank, 1989 ; Leake, 1996) : il est composé d'une *base de cas* constituée d'éléments factuels sur les exploitations enquêtées, d'une *base de connaissances* sur le domaine et d'un *module de raisonnement* à partir de cas. Un *cas* s'entend généralement comme un problème résolu, qui est utilisé dans la résolution d'un nouveau problème (Lieber et Napoli, 1999). Dans le cadre du projet RoSA, un *cas* est une exploitation enquêtée, considérée sous ses aspects d'organisation, spatiale et fonctionnelle. Les objectifs du module de raisonnement sont de comparer et d'adapter les connaissances liées à un cas (par exemple une exploitation *source*, dont on connaît à la fois la structure spatiale et le fonctionnement) à un autre cas (une exploitation *cible* dont on ne connaît que la structure spatiale). La comparaison est établie sur des mesures de similarités entre les structures spatiales des deux exploitations et l'adaptation permet de proposer une logique de fonctionnement à l'exploitation *cible* à partir de la logique de fonctionnement de l'exploitation *source* en s'appuyant sur la similarité entre les structures spatiales. L'hypothèse sous-jacente peut s'exprimer ainsi : *le fonctionnement de* source *est au fonctionnement de* cible *ce que la structure spatiale de* source *est à la structure spatiale de* cible (Lieber, 1997). L'objectif du système RoSA est à la fois un objectif de capitalisation et d'explicitation des connaissances agronomiques. Il doit permettre en effet d'enregistrer des cas (donc des informations sur les exploitations enquêtées) et de mettre en œuvre une forme de raisonnement permettant leur analyse et leur réutilisation pour les enquêtes ultérieures. Ce système ne se veut pas un système automatique, mais un système interactif, d'aide à l'analyse et l'interprétation d'organisations spatiales agricoles.

---

[4] RoSA pour Raisonnement sur des Organisations Spatiales Agricoles.

Le développement du système R<small>OSA</small> s'appuie sur et s'accompagne d'un processus d'acquisition/modélisation de connaissances impliquant agronomes et informaticiens, et confrontant leurs outils et méthodes de représentation de l'espace. Les agronomes utilisent les chorèmes, modèles graphiques de l'organisation de l'espace (Brunet, 1986 ; Cheylan *et al.*, 1990), tandis que les informaticiens ont fait le choix d'utiliser des graphes étiquetés comme support de l'acquisition de connaissances (Le Ber *et al.*, 2003). L'intérêt des graphes réside en effet principalement dans la modalité graphique à la base de la représentation (par opposition au système de représentation et de raisonnement sous-jacent), modalité qui permet de représenter naturellement des structures spatiales. L'outil graphe est par conséquent un outil de modélisation que peuvent ainsi s'approprier facilement à la fois les agronomes et les informaticiens. Dans le cadre de notre projet, nous avons utilisé un formalisme inspiré des graphes conceptuels (Sowa, 1984 ; Chein et Mugnier, 1992). Un graphe est composé de deux ensembles de sommets, l'un représente des entités, l'autre des relations. Les sommets-entités et les sommets-relations peuvent être regroupés en concepts auxquels sont associés des attributs. Les arêtes du graphe sont étiquetées par le rôle[5] des entités dans la relation. Nous utilisons les graphes pour modéliser les chorèmes en supposant que :

- les objets du chorème peuvent être modélisés par les sommets-entités d'un graphe,
- les caractéristiques de ces objets (formes et légendes associées) peuvent être décrites par des attributs associés aux concepts,
- l'arrangement des objets du chorème (leurs relations spatiales) peut être représenté par les sommets-relations d'un graphe.

Afin que le lecteur se rende compte concrètement de la nature d'un graphe, nous présentons ci-dessous (figure 1) un écran du logiciel construit, où l'on voit un graphe modélisant la structure spatiale d'une exploitation caussenarde.

Par ailleurs – et cela sous-tend l'ensemble de la modélisation et du processus d'acquisition de connaissances que nous décrivons ici – nous nous plaçons dans le cadre d'une représentation hiérarchique des connaissances (Ducournau *et al.*, 1998, chapitres 10, 11, 12) : les concepts sont organisés selon une relation d'ordre, fondée sur une classification naturelle du domaine (par exemple *« les parcs sont des surfaces en herbe »*, *« la bergerie est un bâtiment d'exploitation »*, *« un puits est un aménagement »*). Cette relation d'ordre est utilisée par établir une mesure de ressemblance entre les concepts et entre les graphes, et c'est donc sur elle que reposent les opérations de comparaison et d'adaptation du système ROSA. Nous verrons qu'au cours du processus d'acquisition de connaissances cette relation d'ordre sera remise en cause à de nombreuses reprises par les acteurs, agronomes et informaticiens.

---

[5] Ici *rôle* est pris dans le sens des logiques des descriptions (Baader *et al.*, 2003), c'est-à-dire comme une relation entre deux concepts, ici entre un concept-relation et un concept-entité.

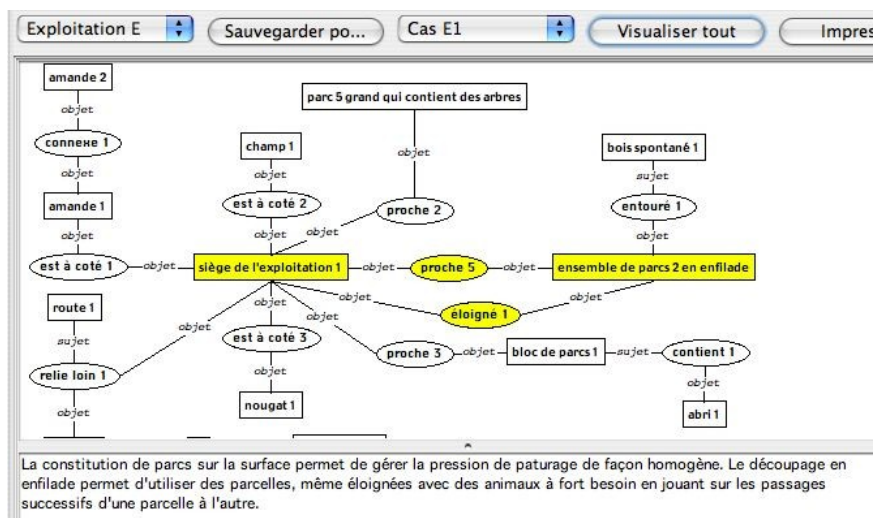

Figure 1 : Interface du logiciel ROSA - Vue d'un graphe décrivant la structure d'une exploitation. Les sommets-entités (par exemple *champ 1*) sont entourés d'un rectangle, les sommets-relations (par exemple *est à côté 2*) d'une ellipse. Les arcs sont étiquetés par un rôle (*objet*, *sujet*). Le sous-graphe coloré en jaune correspond à un cas, c'est-à-dire que cette structure a une fonction particulière dans l'exploitation, explicitée en bas de l'écran.

# PREMIÈRE ÉTAPE : MODÉLISATION DU DOMAINE ET DÉFINITION DES CAS

Comme nous l'avons déjà indiqué plus haut, à la première étape du projet, il s'agit pour l'ensemble des chercheurs du projet de cerner les concepts que les agronomes utilisent pour décrire la structure spatiale et fonctionnelle d'une exploitation agricole. Cette description est en effet un réquisit pour formaliser les 'cas'. À cet effet, nous avons besoin de bien appréhender leur mode de description. Les agronomes du projet utilisent la méthode des chorèmes (Brunet, 1986 ; Cheylan *et al.*, 1990) pour représenter l'espace agricole et élaborer des chorèmes d'exploitation, appelés aussi par la suite schémas chorématiques (Lardon *et al.*, 2000). C'est la raison pour laquelle, la procédure d'acquisition de connaissances s'appuie d'une part sur les chorèmes d'exploitations, utilisés par les agronomes, et d'autre part sur les graphes, utilisés par les informaticiens. Cette procédure d'acquisition se réalise au long d'interactions mettant en scène ces agronomes et informaticiens. Les agronomes mobilisent des chorèmes qu'ils ont élaborés en amont de l'interaction, après qu'ils ont enquêté dans les exploitations concernées. Les informaticiens transforment ces chorèmes en graphes, 'traduisant' ainsi les contenus habitant les premiers en ces représentations graphiques qui pourront faire l'objet d'implantation informatique. L'ensemble se réalise, au vu et au su de tous, au cœur de l'articulation entre l'explicitation des chorèmes par les agronomes et leur traduction en graphes par les informaticiens. Une interrogation systématique du sens porté à la fois par les schémas chorématiques et par la hiérarchisation de leur contenu conduit le groupe à deux choses : l'explicitation des concepts utilisés par les agronomes et la définition de 'cas'.

- La mise au jour des concepts utilisés par les agronomes se réalise d'une manière variée ; elle peut, par exemple, apparaître comme une découverte de concepts utilisés mais non nommés :

```
(2001(7), 39'08)[6]
LAU : alors là je sais pas comment tu vas appeler ca
VIV : ben je sais pas
LAU : des parcs
VIV : englobant des .. des champs
LAU : mmm
VIV : mais c'est
LAU : c'est chaque parc individuellement qui contient des champs
VIV : ouais
THO : mmm
VIV : donc c'est pas un ensemble de parcs
LAU : oui
THO : c'est un ensemble de parcs contenant champs
VIV : et voilà et ca y est on a un concept
```

- La définition de 'cas' porte sur des fragments d'espace dotés d'une organisation et d'une signification particulières :

```
(2001(1), 43'17)
VIV : là c'est bien l'idée de euh ben y'a une contrainte quand
      même je présume que la bordure du ruisseau c'est plus
      humide et cætera
BER : ben voilà
VIV : et donc c'est mais c'est au travers d'une d'une pratique
      de mise en herbe qu'il qui résout son problème
```

Cette activité d'acquisition s'est déroulée au cours de plusieurs séances réunissant un (ou plusieurs) agronome(s) et un (ou plusieurs) informaticien(s) impliqués dans le projet. L'une d'entre elles (en janvier 2001) a fait l'objet d'un filmage. Dans cette séance, la procédure d'acquisition se réalisa au long d'une interaction entre quatre acteurs, deux agronomes et deux informaticiens. Le filmage permet d'avoir accès à l'évènement interactionnel qui s'est déployé dans un espace-temps donné (un studio d'enregistrement et cinq heures de travail). Ainsi et la documentation pré-existante au travail en commun (principalement les parcellaires et les schémas chorématiques) et les formes graphiques et textuelles produites dans cet espace-temps sont-elles disponibles pour analyse. Plus, on dispose de l'histoire de la production des multiples traces laissées par les acteurs. On peut par exemple constater les modifications concrètes (sur-traçages ou biffages) apportées à tel ou tel schéma chorèmatique, ou suivre de façon très précise le processus d'élaboration de tel ou tel graphe, ce suivi mettant à jour hésitations, gommages, raturages ou autres altérations de tel ou tel tracé (pour une présentation et une analyse de ce corpus voir par exemple Brassac et Le Ber, 2005). Bien évidemment l'intégralité des énoncés produits par les acteurs est soumise à transcription et les actions corporelles (jeux de regards, pointage, posture) sont disponibles.

---

[6] Les extraits de corpus mettent en scène trois informaticiens (BER, LAU et JAC) et trois agronomes (PAL, THO, VIV).

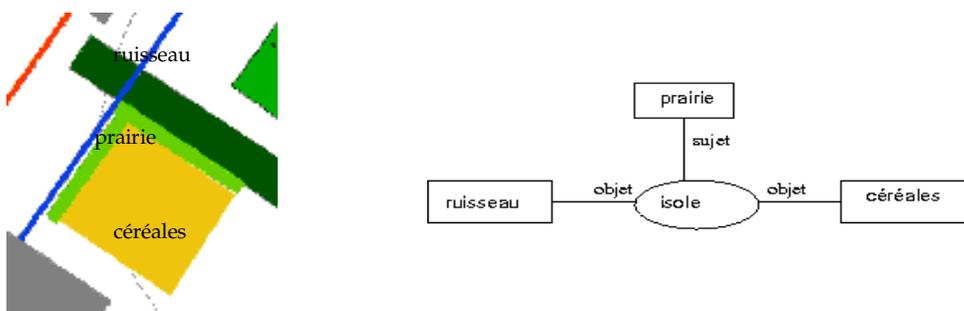

Figure 2: Partie de chorème et sous-graphe associé. Le cas obtenu est la réunion de ce sous-graphe avec l'explication suivante : "l'agriculteur a placé une prairie pour isoler la parcelle de céréales du ruisseau afin de protéger les cultures de l'humidité".

Le travail a porté de façon exhaustive sur sept exploitations, qui avaient donc été 'chorémisées' en amont de cette étape, donnant lieu à environ soixante-dix cas[7] et à la construction d'une hiérarchie de concepts et de relations. De fait, du point de vue de la construction du système, la traduction des chorèmes en graphes se décline en trois étapes :

- écriture des graphes : dénomination des sommets-entités, définition des sommets-relations, fixation des arêtes et des rôles entre relations et entités,

- définition des concepts du domaine : catégorisation des sommets-entités et des sommets-relations, définition des attributs, hiérarchisation des concepts,

---

[7] Chaque cas correspond à un morceau du territoire de l'exploitation ayant une logique fonctionnelle localement propre.

- constitution des cas : explicitation de certaines structures, acquisition d'éléments explicatifs sur le fonctionnement des exploitations agricoles.

À l'issue de cette première étape, le système Rosa se trouve donc doté d'une base de cas et d'une base de connaissances. La base de cas contient des graphes représentant des chorèmes d'exploitations. Chaque graphe peut être découpé en sous-graphes et donner ainsi lieu à plusieurs cas tel que celui présenté dans l'illustration 2. Pour être tout à fait clair, un cas au sens strict est une paire (sous-graphe, explication) ; on pourra également qualifier de cas, au sens plus large, un triplet (portion de chorème, sous-graphe, explication). La base de connaissances contient les concepts (entités spatiales et relations) du domaine. L'une et l'autre ont été construites au moyen de la traduction des chorèmes en graphes et du questionnement des connaissances des agronomes qui en découle. La combinaison des deux va permettre de mener à bien les opérations de comparaison et d'adaptation pour les exploitations qui seront étudiées dans la suite du projet.

# DEUXIÈME ÉTAPE : RECHERCHE DES CONNAISSANCES D'ADAPTATION

La base de cas est dorénavant disponible. Pour les sept exploitations initialement enquêtées, les chercheurs disposent de deux types de formalisations, l'une graphique au rapport figuratif assez étroit au terrain, les chorèmes, l'autre, elle aussi graphique, mais dont le rapport figuratif au terrain est beaucoup plus lâche, les graphes. Ces derniers sont directement codables et peuvent donc faire l'objet d'un traitement numérique. L'idée est de pouvoir, pour d'autres exploitations, raisonner à partir des cas mis à jour lors de la constitution de la base de cas. La deuxième étape consiste donc à examiner d'autres exploitations, chorémisées, et à tenter d'utiliser le rapport structure spatiale/modalité de fonctionnement, formalisé pour les sept premières exploitations sources (es), pour toutes les exploitations cibles (ec) qui pourraient profiter de ce travail. C'est l'objet de cette deuxième étape.

Cette utilisation de l'acquis passe par la comparaison entre un fragment d'une 'ec' avec l'un des fragments des 'es'. La procédure mise en place consiste à comparer des chorèmes et/ou des graphes de différentes exploitations. Cela nécessite la mobilisation de connaissances d'ordre agronomiques et informatiques. Elle implique donc agronomes et informaticiens. L'évaluation des fragments d'exploitations conduit à des mesures de ressemblance et dissemblance entre des structures spatiales et fonctionnelles. Elle provoque également la mise au jour de connaissances relatives au rapport structure spatiale – fonctionnement des exploitations ; connaissances dites d'adaptation car elles étayent les arguments des uns et des autres pour justifier tel ou tel rapprochement entre une 'es' et une 'ec', elles légitiment le fait d'utiliser l'explication associée à une 'ec' pour interpréter la structure spatiale d'une 'es'.

Là encore, ce travail s'est déroulé en plusieurs séances réunissant ces deux groupes de chercheurs. L'une d'elles, mise en place en mars 2003, a donné lieu à une captation vidéographique. Lors de cette séance, les auteurs des propositions d'appariement changent :

- ce sont quelquefois les agronomes qui propose un rapprochement entre fragments (telle situation dans une exploitation *est comparable* ou non à telle autre)

```
(2003(11), 47')
PAL : non alors là je je pen- je je je je pense et je suis pas tout à
      fait sûr mais je pense justement là dans ce cas de figure de de
      de de Soulage y a pas la figure du nougat de l'amande
```

- ce sont quelquefois les informaticiens qui le font (tel schéma chorématique *ressemble* à tel autre, tel graphe *s'apparie simplement* avec tel autre).

```
(2003(22), 12'34 avec des coupures)
JAC : non je demande si ça ressemble
VIV : y a un parc qui contient des du bois et ... et qui ... a un accès
      euh ... vers le parc ... deux
JAC : là j'ai un parc qui contient du bois et y a
VIV : et a un accès sur le parc ... trois
JAC : d'accord
VIV : oui ça j ça ressemble
JAC : par rappo et par rapport à cette solution ... par rapport à cette
      explication
```

La séance s'est déroulée en deux temps. Le premier temps fut consacré à la comparaison de chorèmes d'exploitations sur deux zones différentes (Causse Méjan, Causse Sauveterre), le deuxième temps à l'appariement (manuel) de graphes : l'agronome produit un graphe d'exploitation à partir d'un chorème, puis l'informaticien(ne) cherche à apparier des parties du graphe créé avec les graphes des cas (sources) déjà constitués. L'agronome valide ou non l'appariement, puis évalue et adapte éventuellement l'explication associée au cas source considéré.

Ce travail résulte en différents points concernant le système Rosa :

- une validation des cas (sources) décrits dans la base de cas : ils sont vérifiés et utilisés
- un affinage du modèle du domaine avec introduction de nouveaux concepts (par différentiation) et remise en cause partielle de sa structure hiérarchique

```
(2003(11), 41'13)
VIV : le pacage cc'qu'on c'est le terme qu'on a utilisé pour parler de
      ces surfaces. En herbe. De type parcours. Ou parc. Mais euh qui
      qu'avaient été améliorés donc euh qu'étaient labourables
LAU : donc le pacage ça peut être du parc ou du parcours
PAL : non pour moi j'pense qui qu'ils sont tous parc oui
```

- la mise en évidence de modèles généraux d'organisation du territoire, qui peuvent servir à organiser les cas et mener à des « structures-concepts ».

```
(2003(22), 14'38)
PAL :l'idée c'est qu'une surface donne accès à une autre surface
      commodément à partir d'une bergerie
(15'14) alors que jusqu'à présent on était sur des chemins ou sur des
      proximités des proximités des chemins des rues des drailles des
      tout ce qu'on veut
```

À l'issue de cette seconde étape de l'étude, le groupe de chercheurs est maintenant capable de mettre en œuvre des procédures d'appariement qui sont à la base du module de *raisonnement à partir de cas* dont nous avons parlé plus haut. Au-delà de cette mise en œuvre, les informaticiens sont en mesure de construire ce module. La machinerie informationnelle, qui constituera le cœur de l'outil d'aide à la décision visé par le groupe, est ainsi opérationnelle[8]. Elle se présente sous la forme d'un prototype qui a évidemment besoin d'une validation. C'est cette procédure de validation qui constitue la troisième étape du processus.

# TROISIÈME ÉTAPE : ÉVALUATION DU PROTOTYPE

Les deux premières étapes ont conduit le groupe à l'état suivant. Une base de cas et une base de connaissances ont été élaborées ; le module de raisonnement à partir de cas est opérationnel. Plus exactement les procédures d'appariement de graphes sont implantées et prêtes à être évaluées. Cette évaluation va se réaliser en deux séances (août 2004 et février 2005), la seconde seulement étant filmée. Celle-ci est lourdement instrumentée. Le prototype est là 'dans' un ordinateur qu'un des deux informaticiens manipule tout au long des échanges. L'écran de cet ordinateur est projeté en temps réel sur un tableau blanc (effaçable) *via* un vidéoprojecteur, les acteurs pouvant en continu venir surligner, cercler, barrer ce qui est projeté et également écrire ou faire un schéma à côté des éléments projetés, comme on le voit dans l'illustration ci-dessous (figure 3).

---

8 Rosa utilise pour cela le système Racer, qui est fondé sur une logique de descriptions et permet la classification de concepts (Haarslev *et al.*, 2001).

Dans l'état présent du prototype, l'usage du module de raisonnement permet des appariements entre un graphe cible et plusieurs cas source. Par exemple, lorsque le groupe veut travailler sur une exploitation 'nouvelle', ses membres examinent le chorème, élaborent un graphe associé et soumettent au module telle ou telle partie du graphe. Le module, après calcul, fournit des cas (graphe plus explication) issus de la base de cas, qui présentent des similitudes hiérarchisées avec le cas cible. La tâche alors effectuée par les membres du groupe est d'évaluer ces similitudes. Le test du prototype consiste en cette ratification/invalidation des appariements proposés. Ces acceptations et rejets sont bien sûr argumentés par les agronomes et/ou informaticiens.

On voit là que le travail réalisé par le groupe au cours même des discussions de cette séance est très proche de ce qu'ils ont fait en 2003. La différence est que cette fois-ci, les échanges sont préparés par l'implantation. En effet, l'appariement est le fait d'une automatisation elle-même résultat d'une numérisation des connaissances produites antérieurement par les membres du groupe. De plus, le travail conduit également à l'évaluation de l'ergonomie du logiciel et de son utilisabilité par les agronomes.

Il faut signaler que cette troisième étape, étape du test du prototype construit, donne lieu à une réinterrogation des connaissances construites lors des deux premières phases. En effet, les tests d'utilisation du prototype Rosa permettent à la fois d'évaluer :

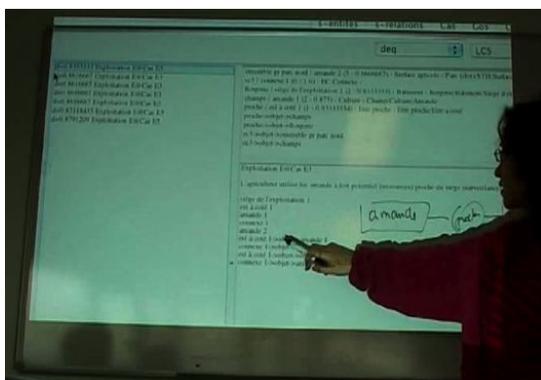

Figure 3 : Séance 2005 - visualisation et discussion sur écran blanc des cas cible et source appariés par le logiciel Rosa.

- le contenu de la base de cas : les graphes et leur explication associée. Certains graphes sont modifiés (suppression, adjonction ou changement d'un sommet), certaines explications réécrites.

- la structure de la base de connaissances : les concepts et leur organisation hiérarchique. En particulier, au-delà de la simple représentation hiérarchique des concepts, on voit qu'il est nécessaire de représenter des continuums.

```
(2005(V2), 20'50)
VIV : l'idée c'est que à la fois bloc nougat euh fertilisé petit proche
      qui fait que la surface en herbe bascule plutôt du côté des des
      des bonnes surfaces typ- des type champ
```

- le mode opératoire : méthode d'appariement entre graphes, appariement autorisés entre concepts.

```
(2005(V1), 1h22'00)
LAU : et est ce que les amandes on peut les associer aux parcs
VIV : oui
LAU : est ce que les
VIV : NON aux champs
LAU : aux champs. Les amandes on peut les associer aux champs. Et pas
      aux parcs. Même à des petits parcs
VIV : dans certains cas les petits parcs euh bien bien fertiles
PAL : les bons petits parcs soignés si
```

Ces dernières observations n'ont pas donné lieu à une évolution concrète du prototype, qui est resté en l'état à l'issue du projet. Différentes voies d'amélioration sont encore à l'étude.

## SYNTHÈSE ET CONCLUSION

Le projet s'est construit autour de trois questions-propositions émanant des agronomes, des informaticiens et des psychologues et linguistes. Les premiers s'interrogeaient sur les outils de représentation de l'espace agricole et sur leurs capacités d'explicitation et de généralisation des modes de gestion des agriculteurs. Les seconds s'intéressaient à la modélisation de raisonnements sur l'espace, proposant de développer des approches qualitatives aptes à rendre compte de la singularité et de la complexité de ces modes de gestion. Les troisièmes, quant à eux, étudiaient la mobilisation d'objets dans la construction et le développement d'une interaction sociale finalisée et se sont penchés – dans une vision réflexive – sur la confrontation des modèles agronomiques et informatiques.

La mise en place des trois expérimentations – réunissant les trois regards – s'est faite de façon progressive, en fonction des avancées du travail de développement du logiciel RosA (Metzger, 2005). Les trois expérimentations correspondent en effet à trois phases-clé de ce développement, qui a duré de fin 2000 à 2005 : la première phase concerne la modélisation des connaissances du domaine, la deuxième la modélisation du raisonnement, et la troisième le test du logiciel. Les trois expérimentations ont résulté en trois corpus vidéos qui ont été diversement exploités (Capitaine *et al.*, 2001 ; Le Ber *et al*., 2002 ; Mondada, 2005) mais qui ont tous donné lieu à des confrontations et des apports concrets entre les disciplines concernées, comme le montrent les différents textes qui constituent ce dossier. Cet entrelacement des intérêts, des méthodes et des concepts place ce 'chantier' de recherche, intrinsèquement interdisciplinaire et délibérément concret (il a donné lieu à un développement informatique opérant), dans le champ d'une anthropologie des connaissances, ici relatives à la gestion des territoires agricoles.